\documentclass{article} 
\usepackage{iclr2026_conference,times}

\usepackage{amsmath,amsfonts,bm}

\def\eqref#1{equation~\ref{#1}}

\def\1{\bm{1}}

\DeclareMathAlphabet{\mathsfit}{\encodingdefault}{\sfdefault}{m}{sl}
\SetMathAlphabet{\mathsfit}{bold}{\encodingdefault}{\sfdefault}{bx}{n}

\usepackage{hyperref}
\usepackage{url}
\usepackage{graphicx}
\usepackage{booktabs} 
\usepackage{multirow} 
\usepackage{float}
\usepackage{tcolorbox}

\tcbuselibrary{breakable,skins,listings}
\usepackage{listings}

\lstdefinestyle{promptcode}{
  basicstyle=\ttfamily\small,
  breaklines=true,
  breakatwhitespace=false,
  columns=fullflexible,
  keepspaces=true,
  showstringspaces=false
}

\title{Diagnosing Retrieval vs. Utilization Bottlenecks in LLM Agent Memory}

\author{
Boqin Yuan \\
University of California, San Diego \\
\texttt{b4yuan@ucsd.edu}
\And
Yue Su \\
Carnegie Mellon University \\
\texttt{yuesu@andrew.cmu.edu}
\And
Kun Yao \\
University of North Carolina \\
\texttt{kya@unc.edu}
}

\iclrfinalcopy 
\begin{document}

\maketitle

\begin{abstract}
Memory-augmented LLM agents store and retrieve information from prior interactions, yet the relative importance of \emph{how} memories are written versus \emph{how} they are retrieved remains unclear. We introduce a diagnostic framework that analyzes how performance differences manifest across write strategies, retrieval methods, and memory utilization behavior, and apply it to a $3 \times 3$ study crossing three write strategies (raw chunks, Mem0-style fact extraction, MemGPT-style summarization) with three retrieval methods (cosine, BM25, hybrid reranking). On LoCoMo, retrieval method is the dominant factor: average accuracy spans 20 points across retrieval methods (57.1\% to 77.2\%) but only 3--8 points across write strategies. Raw chunked storage, which requires zero LLM calls, matches or outperforms expensive lossy alternatives, suggesting that current memory pipelines may discard useful context that downstream retrieval mechanisms fail to compensate for. Failure analysis shows that performance breakdowns most often manifest at the retrieval stage rather than at utilization. We argue that, under current retrieval practices, improving retrieval quality yields larger gains than increasing write-time sophistication.
\end{abstract}

\section{Introduction}
A growing body of work equips LLM agents with persistent memory~\citep{packer2023memgpt, xu2025amem, chhikara2025mem0}. These systems differ mainly in what they store: some keep raw conversation text~\citep{lewis2020rag}, others extract structured facts with conflict resolution~\citep{chhikara2025mem0, xu2025amem}, and still others compress sessions into summaries~\citep{packer2023memgpt}. Recent work has pushed further with inter-memory linking~\citep{xu2025amem} and reinforcement learning for end-to-end memory management~\citep{zhou2025mem1learningsynergizememory, wang2025memalphalearningmemoryconstruction}. Yet a basic question remains open: \emph{does the write strategy actually matter for downstream performance, or do performance differences primarily arise at retrieval?} Current benchmarks~\citep{maharana2024locomo, wu2025longmemeval} only measure end-to-end accuracy, making it hard to tell whether errors stem from what was stored, how it was retrieved, or how the LLM used the context (Figure~\ref{fig:pipeline}). We review related work in Appendix~\ref{app:related-work}.

We address this with two contributions. First, we propose a \textbf{diagnostic probing framework} that sits at the retrieval-to-generation boundary and independently measures retrieval relevance, memory utilization, and failure modes. Second, we conduct a \textbf{controlled $3 \times 3$ factorial study} that crosses three write strategies (raw chunks, Mem0-style extraction, MemGPT-style summarization) with three retrieval methods (cosine similarity, BM25, hybrid reranking); see Appendix~\ref{app:experiment-design} for a detailed discussion of the design rationale. Evaluating on LoCoMo (1,540 non-adversarial questions), we find that retrieval method is the dominant factor: accuracy varies by 20 points across retrieval methods but only 3 to 8 points across write strategies. The cheapest write approach, storing raw chunks with zero LLM calls, matches or outperforms the more expensive alternatives.

\begin{figure}[t]
  \centering
  \begin{minipage}[t]{0.45\linewidth}
    \centering
    \includegraphics[height=4.5cm]{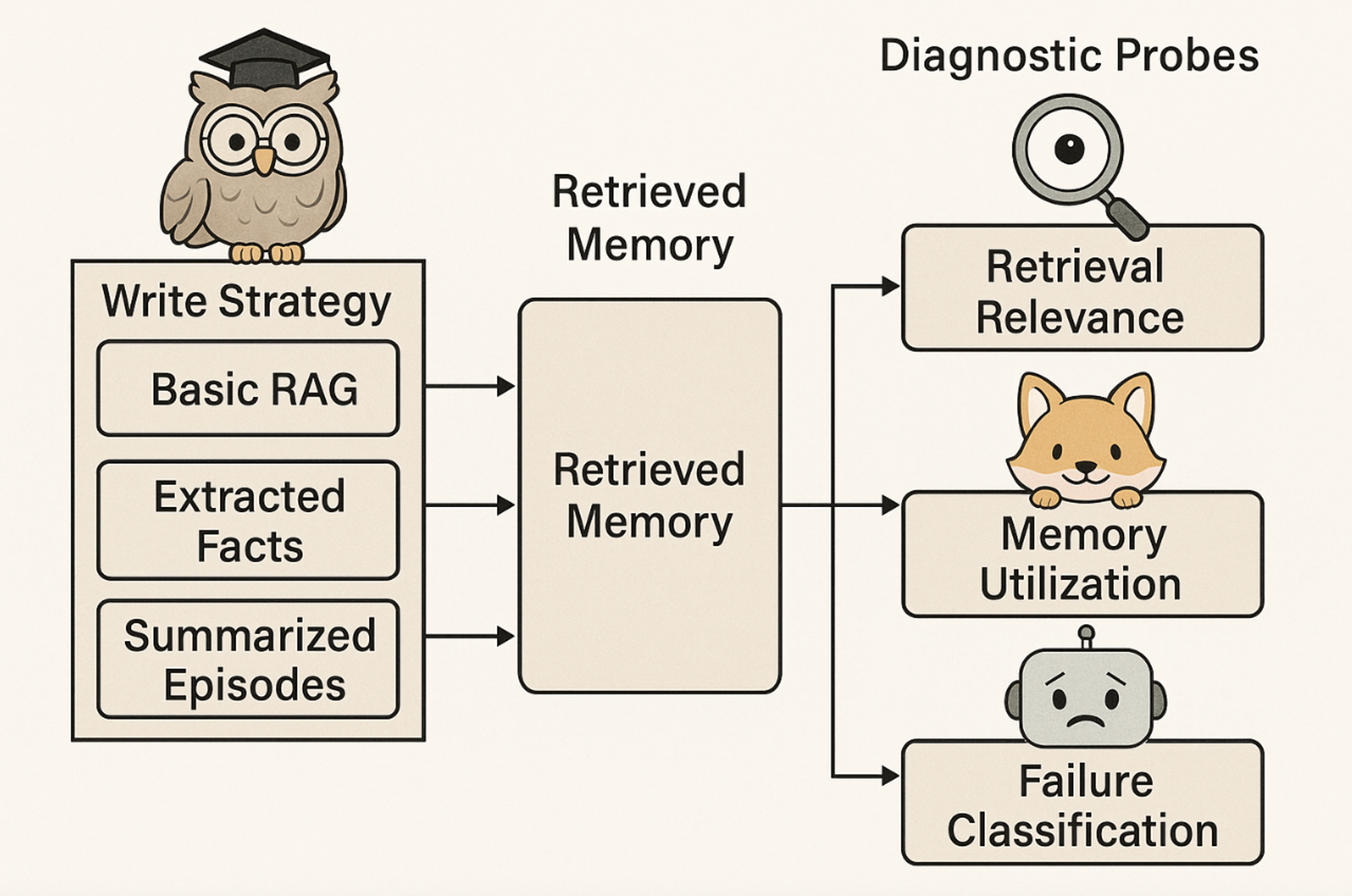}
    \caption{Memory-agent pipeline with three diagnostic probes at the retrieval-to-generation boundary.}
    \label{fig:pipeline}
  \end{minipage}
  \hfill
  \begin{minipage}[t]{0.52\linewidth}
    \centering
    \includegraphics[height=4.5cm]{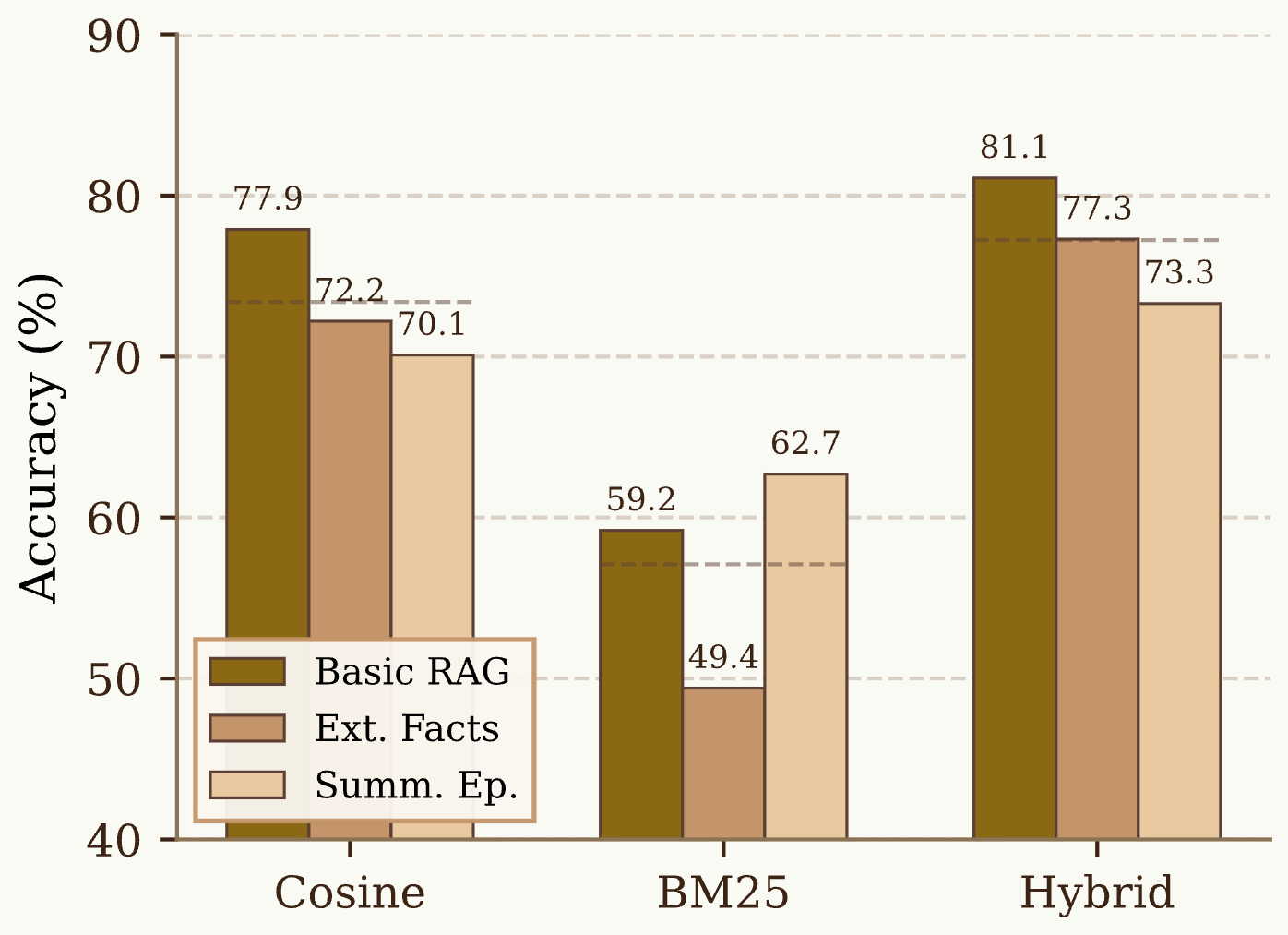}
    \caption{Accuracy across the $3 \times 3$ grid ($k{=}5$). Retrieval method drives 14--23 point differences; write strategy only 3--8.}
    \label{fig:main-results}
  \end{minipage}
\end{figure}

\section{Method}

\subsection{Write Strategies}
\label{sec:write}

We evaluate three prompt-based write strategies on LoCoMo~\citep{maharana2024locomo} (10 multi-session conversations, $\sim$600 turns each, $\sim$200 questions per conversation). All configurations use GPT-5-mini as the backbone LLM and \texttt{text-embedding-3-small} (1536-d) for embeddings.

\textbf{Basic RAG} stores raw 3-turn conversation chunks with speaker names and timestamps, requiring zero LLM calls at write time.  
\textbf{Extracted Facts} follows Mem0~\citep{chhikara2025mem0}: an LLM extracts self-contained facts per session, followed by embedding-based matching and LLM conflict resolution (\textsc{add}/\textsc{update}/\textsc{noop}).  
\textbf{Summarized Episodes} follows MemGPT~\citep{packer2023memgpt}: each session is compressed into a single summary paragraph. Table~\ref{tab:design} summarizes computational costs; full prompts and implementation details are provided in Appendix~\ref{app:memory-prompts}.

\subsection{Retrieval Methods}
\label{sec:retrieval}

We pair each write strategy with three retrieval methods of increasing sophistication, with a default retrieval budget of $k{=}5$. \textbf{Cosine similarity} returns the top-$k$ memory entries by cosine distance between the query embedding and each stored entry's embedding; this is the default method in most deployed memory systems~\citep{chhikara2025mem0, xu2025amem} and captures semantic similarity, but may miss lexically relevant entries with different surface forms. \textbf{BM25} scores entries by term-frequency-based keyword overlap, complementing embedding-based retrieval by surfacing entries that share exact terms with the query; however, it fails when relevant memories use different vocabulary than the question. \textbf{Hybrid+Rerank} first pools the top-$2k$ candidates from both cosine and BM25, then uses GPT-5.2 as an LLM judge to rerank the union down to top-$k$; this adds one LLM call per query but leverages both semantic and lexical signals, with the reranker resolving disagreements between the two retrieval sources.

\subsection{Probing Framework}

For each question $q$ with gold answer $a^*$, we generate an answer $a_\text{mem}$ \emph{with} retrieved memory (top-$k$ entries in the prompt) and an answer $a_\text{no}$ \emph{without} any memory, using the same LLM and prompt template. We then apply three diagnostic probes. \textbf{Probe 1 (Retrieval Relevance):} An LLM judge evaluates whether each retrieved entry $m_i$ contains information relevant to answering $q$; we report Retrieval Precision@$k$ = $|\{m_i : \text{relevant}\}| / k$. \textbf{Probe 2 (Memory Utilization):} An LLM judge compares $a_\text{mem}$ and $a_\text{no}$ against $a^*$ and classifies each question as \textbf{Beneficial} (memory improved the answer), \textbf{Harmful} (memory worsened it), \textbf{Ignored} (answer unchanged), or \textbf{Neutral} (answer changed but correctness unaffected). \textbf{Probe 3 (Failure Classification):} For incorrect answers, we group failures into three categories: \emph{Retrieval failure} covers cases where the system did not surface sufficient information to answer the question at inference time. This includes (i) cases where relevant information was not retrieved despite being present in the memory store, and (ii) cases where the stored memories did not contain sufficient detail to support the answer. We group both under retrieval-stage failure because the breakdown manifests at the retrieval-to-generation boundary; \emph{Utilization failure} applies when at least one relevant memory was retrieved but the model still produced an incorrect answer, indicating a downstream reasoning problem; and \emph{Hallucination} captures the rare cases where the model's answer directly contradicts the content of its own retrieved memories.

\section{Results}

\subsection{Write Strategy Has Minimal Impact}
Table~\ref{tab:main-grid} shows that write strategy accounts for only 3--8 accuracy points within any retrieval column. Basic RAG, which stores raw chunks with zero LLM calls, achieves 77.9\% under cosine and 81.1\% under hybrid, matching or exceeding both Extracted Facts and Summarized Episodes across all retrieval methods. The one exception is BM25, where Summarized Episodes (62.7\%) edges out Basic RAG (59.2\%), likely because keyword retrieval benefits from the denser vocabulary of compressed summaries. Overall, the more expensive write strategies do not justify their cost: lossy compression discards conversational details that the backbone LLM can otherwise leverage directly (Figure~\ref{fig:main-results}).

\begin{table}[H]
\caption{Token F1 and LLM-as-Judge accuracy across all nine configurations ($k{=}5$, 1,540 non-adversarial questions). Accuracy spread across retrieval methods (14--23 pts) consistently exceeds spread across write strategies (3--8 pts). Bold = best write strategy per retrieval method.}
\label{tab:main-grid}
\begin{center}
\small
\begin{tabular}{lccc|ccc}
\toprule
& \multicolumn{3}{c}{\textbf{Token F1}} & \multicolumn{3}{c}{\textbf{Accuracy (\%)}} \\
\cmidrule(lr){2-4} \cmidrule(lr){5-7}
& \textbf{Cos} & \textbf{BM25} & \textbf{Hyb} & \textbf{Cos} & \textbf{BM25} & \textbf{Hyb} \\
\midrule
Basic RAG & \textbf{0.232} & \textbf{0.184} & \textbf{0.240} & \textbf{77.9} & 59.2 & \textbf{81.1} \\
Extracted Facts & 0.211 & 0.146 & 0.220 & 72.2 & 49.4 & 77.3 \\
Summ.\ Episodes & 0.197 & 0.173 & 0.202 & 70.1 & \textbf{62.7} & 73.3 \\
\midrule
Avg & 0.213 & 0.168 & 0.221 & 73.4 & 57.1 & 77.2 \\
\bottomrule
\end{tabular}
\end{center}
\end{table}

\subsection{Retrieval Method Dominates}
\begin{figure}[t]
  \centering
  \begin{minipage}[t]{0.48\linewidth}
    \centering
    \includegraphics[height=4.2cm]{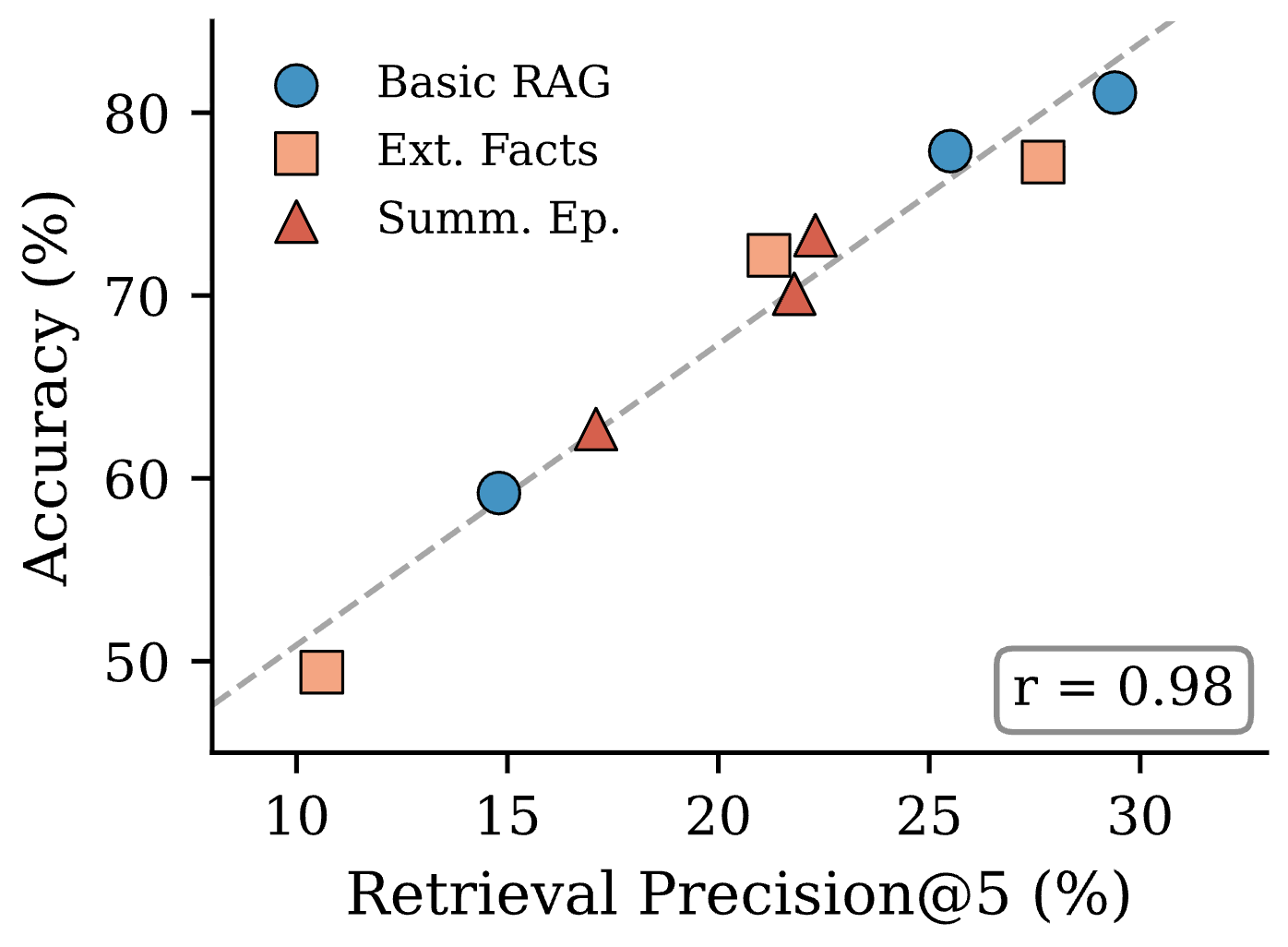}
    \caption{Precision@5 vs. accuracy ($r{=}0.98$). Each point is one of nine configurations.}
    \label{fig:scatter}

  \end{minipage}
  \hfill
  \begin{minipage}[t]{0.48\linewidth}
    \centering
    \includegraphics[height=4.2cm]{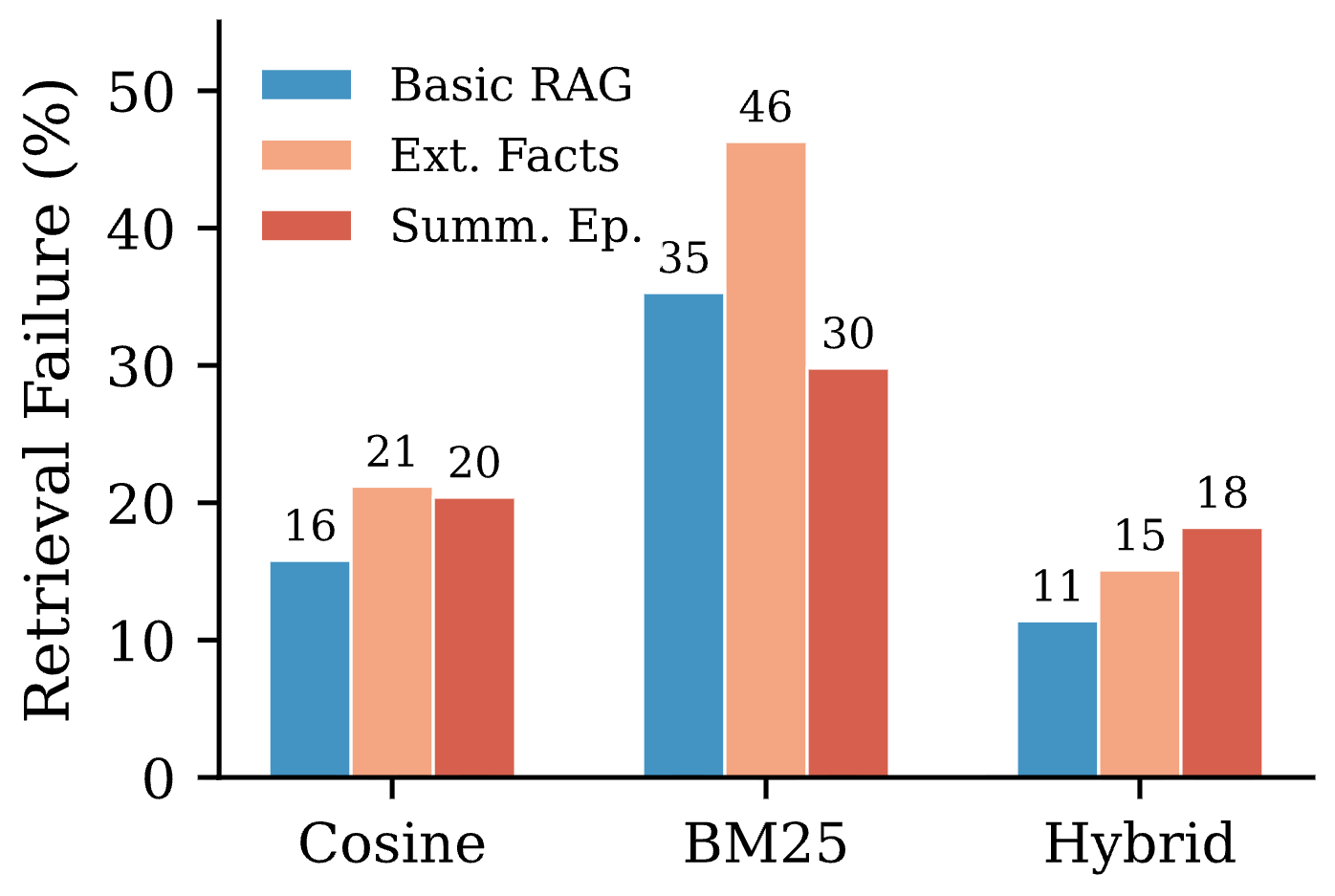}
    \caption{Retrieval failure rate by retrieval method. Hybrid reranking cuts retrieval failures by half or more across all write strategies.}
    \label{fig:failure-bars}
  \end{minipage}
\end{figure}

Switching retrieval method shifts accuracy by 14--23 points, dwarfing the write strategy effect. Hybrid reranking averages 77.2\% across write strategies, compared to 73.4\% for cosine and 57.1\% for BM25. This 20-point gap between hybrid and BM25 holds regardless of what is stored, confirming that surfacing the right context matters far more than how that context was formatted. Retrieval precision is near-perfectly correlated with downstream accuracy ($r{=}0.98$, Figure~\ref{fig:scatter}), providing strong evidence that retrieval quality most strongly correlates with downstream performance. We also note that token F1 is largely insensitive to these accuracy differences (0.221 vs.\ 0.168 for a 20-point accuracy gap), underscoring the importance of LLM-judge evaluation over surface-level overlap metrics.

\subsection{Failure Analysis Confirms Retrieval as the Bottleneck}
Table~\ref{tab:all-probes} and Figure~\ref{fig:failure-bars} break down where failures actually occur. Retrieval failure is the dominant error mode, accounting for 11--46\% of all questions depending on configuration. Under BM25 with Extracted Facts, retrieval failure alone reaches 46.3\%, nearly matching the total error rate. Hybrid reranking cuts retrieval failures roughly in half (e.g., 35.3\% to 11.4\% for Basic RAG). By contrast, utilization failures remain stable at 4--8\% and hallucinations at 0.4--1.4\% regardless of configuration, indicating that the LLM reliably uses relevant context when it is provided. The utilization probe reinforces this: beneficial rates reach 79.0\% under Basic RAG + Hybrid, meaning the model improves its answer four out of five times when given good retrievals. These patterns consistently point to the same conclusion: performance breakdowns most often manifest at the retrieval stage rather than at utilization.

\begin{table}[H]
\caption{Diagnostic probes across all nine configurations (\%). Probe~1: retrieval precision. Probe~2: memory utilization. Probe~3: failure modes as percentage of all questions (remainder = correct). See Section~3.3.}
\label{tab:all-probes}
\centering
\small
\begin{tabular}{llc|cc|ccc}
\toprule
& & \multicolumn{1}{c}{\textbf{Probe 1}} & \multicolumn{2}{c}{\textbf{Probe 2: Utilization}} & \multicolumn{3}{c}{\textbf{Probe 3: Failure Mode}} \\
\cmidrule(lr){3-3} \cmidrule(lr){4-5} \cmidrule(lr){6-8}
& & \textbf{Precision} & \textbf{Beneficial} & \textbf{Ignored} & \textbf{Retrieval} & \textbf{Utilization} & \textbf{Halluc.} \\
\midrule
\multirow{3}{*}{Basic RAG}
& Cosine & 25.5 & 75.7 & 9.2 & 15.8 & 5.4 & 1.0 \\
& BM25 & 14.8 & 56.5 & \textbf{24.8} & 35.3 & 5.1 & 0.4 \\
& Hybrid & \textbf{29.4} & \textbf{79.0} & \textbf{6.6} & \textbf{11.4} & 6.2 & 1.2 \\
\midrule
\multirow{3}{*}{Ext.\ Facts}
& Cosine & 21.2 & 70.4 & 9.7 & 21.2 & 5.9 & 0.6 \\
& BM25 & 10.6 & 46.2 & \textbf{29.2} & \textbf{46.3} & 3.9 & 0.5 \\
& Hybrid & \textbf{27.7} & \textbf{75.3} & 7.2 & 15.1 & 6.9 & 0.7 \\
\midrule
\multirow{3}{*}{Summ.\ Ep.}
& Cosine & 21.8 & 67.9 & 11.4 & 20.4 & 8.1 & 1.4 \\
& BM25 & 17.1 & 60.4 & 18.1 & \textbf{29.8} & 6.4 & 1.0 \\
& Hybrid & \textbf{22.3} & \textbf{70.1} & 9.9 & 18.2 & 7.1 & 1.4 \\
\bottomrule
\end{tabular}
\end{table}

\section{Discussion and Conclusion}

Our results show that retrieval quality has a substantially larger impact on performance than write strategy. Fact extraction and summarization inevitably discard contextual nuances useful at inference time, while raw chunking preserves the full signal with zero LLM calls and matches or outperforms costlier alternatives. Hybrid reranking confirms that most failures stem from irrelevant retrieval rather than memory construction deficiencies; when relevant context is surfaced, the model uses it effectively. More broadly, these findings suggest that modern LLMs already possess strong contextual reasoning, and our results suggest that performance differences are more strongly associated with information selection than with representation choices. This reframes design priorities for memory-augmented agents: progress may depend more on retrieval precision, reranking, and query understanding than on increasingly complex write pipelines.

\paragraph{Limitations.} Our study evaluates a single backbone model (GPT-5-mini) on one benchmark (LoCoMo) with a fixed retrieval budget ($k{=}5$), limiting generality. The write strategies are prompt-based reimplementations rather than fully learned memory systems, and results may differ for reinforcement learning approaches that jointly optimize writing and retrieval. The advantage of raw chunking may also diminish under tighter context budgets where compression is required to fit within model input limits. Additionally, answer correctness and failure classification rely on LLM judges. Extending this analysis across models, datasets, retrieval budgets, and learned memory systems would strengthen the robustness of our conclusions.

\subsubsection*{Acknowledgments}
We thank our annotators for their work on failure classification and answer validation. Experiments were supported by API credits for OpenAI models inference. Portions of the manuscript were edited with assistance from a large language model to improve clarity and presentation. All experimental design, analysis, and conclusions were developed by the authors.

\bibliography{iclr2026_conference}
\bibliographystyle{iclr2026_conference}

\newpage
\appendix

\section{Related Work}
\label{app:related-work}

Memory systems for LLM agents span a design spectrum from raw storage to learned management. MemGPT~\citep{packer2023memgpt} uses recursive summarization; Mem0~\citep{chhikara2025mem0} extracts facts with conflict resolution; A-MEM~\citep{xu2025amem} adds inter-memory linking; and MemoryBank~\citep{zhong2023memorybankenhancinglargelanguage} combines raw logs with hierarchical summaries and forgetting curves. MEM1~\citep{zhou2025mem1learningsynergizememory} and Mem-$\alpha$~\citep{wang2025memalphalearningmemoryconstruction} learn memory management via reinforcement learning, falling outside our prompt-based scope. Several benchmarks evaluate long-term memory, including LoCoMo~\citep{maharana2024locomo}, LongMemEval~\citep{wu2025longmemeval}, MemoryAgentBench~\citep{hu2025evaluatingmemoryllmagents}, and RealTalk~\citep{lee2025realtalk21dayrealworlddataset}, but all measure downstream accuracy without decomposing \emph{why} a system fails. Our probing framework fills this gap.

\section{Experimental Design and Rationale}
\label{app:experiment-design}

\paragraph{Dataset Details.}
LoCoMo contains 10 multi-session conversations (approximately 600 turns each) with roughly 200 questions per conversation spanning single-hop, multi-hop, temporal reasoning, and open-domain categories (adversarial excluded). We evaluate on 1,540 non-adversarial questions.

Memory-augmented agents are often evaluated as end-to-end systems, which obscures where errors arise. A single accuracy number conflates three factors: whether the write strategy preserved the relevant information, whether retrieval surfaced it, and whether the model used it correctly. While our probes separate retrieval-stage breakdowns from utilization errors, we do not independently measure write-time information loss; instead, we analyze how failures manifest at inference time. Our experimental design therefore enables a controlled comparison of write strategies and retrieval methods, allowing us to identify which stage most strongly correlates with downstream performance.

We adopt a $3 \times 3$ factorial design that crosses three write strategies with three retrieval methods (Table~\ref{tab:design}). The write strategies span the information preservation spectrum: Basic RAG stores raw conversation chunks with no processing; Extracted Facts compresses sessions into structured factual statements with conflict resolution; and Summarized Episodes condenses each session into a narrative summary. The retrieval methods span common approaches: cosine similarity for semantic matching, BM25 for lexical matching, and Hybrid+Rerank, which combines both signals with an LLM reranking step. Evaluating all nine combinations allows us to measure the independent effect of writing and retrieval, as well as potential interactions between them.

All configurations use the same backbone model, embedding model, retrieval budget, and benchmark questions to control for implementation variance. We then apply our three diagnostic probes to every setting, enabling analysis of not only overall performance differences but also the specific failure modes driving them.

\begin{table}[H]
\caption{Experimental design. We cross three write strategies (what is stored) with three retrieval methods (how it is retrieved), yielding a $3 \times 3$ evaluation grid.}
\label{tab:design}
\begin{center}
\small
\vspace{0.5\baselineskip}
\begin{tabular}{llll}
\toprule
& \textbf{Component} & \textbf{Mimics / Mechanism} & \textbf{LLM Calls} \\
\midrule
\multirow{3}{*}{\textbf{Write}}
& Basic RAG & Store raw 3-turn chunks & None \\
& Extracted Facts & Mem0-style fact extraction + conflict resolution & 1+/session \\
& Summ.\ Episodes & MemGPT-style session summarization & 1/session \\
\midrule
\multirow{3}{*}{\textbf{Retrieval}}
& Cosine Similarity & Embedding similarity & None \\
& BM25 & Keyword matching & None \\
& Hybrid+Rerank & Cosine $\cup$ BM25, LLM rerank to top-$k$ & 1/query \\
\bottomrule
\end{tabular}
\vspace{0.5\baselineskip}
\end{center}
\end{table}

\section{Memory Prompts}
\label{app:memory-prompts}

\subsection{Fact Extraction (Extracted Facts Strategy)}

\begin{tcolorbox}[
  colback=gray!5!white,
  colframe=gray!75!black,
  title=Extraction Prompt,
  breakable
]
\small\ttfamily\raggedright
You are a memory extraction agent. Given a conversation session between two people, extract ALL important facts, events, preferences, and details mentioned.

For each fact, provide:

- fact: A concise, self-contained statement (should make sense without the original context)
- speakers: Which speaker(s) this fact is about
- type: One of [event, preference, relationship, plan, personal\_detail, opinion]

Conversation:
\{conversation\}

Return a JSON object with key "facts" containing a list of extracted facts.
Example format:
{"facts": 
  [
    {"fact": "Alice adopted a golden retriever named Max in March",
    "speakers": ["Alice"],
    "type": "event"}, 
    ...]
}

Extract every meaningful piece of information. Be thorough, missing a fact means it's lost forever.
\end{tcolorbox}

\vspace{0.3cm}

\subsection{Conflict Resolution (Extracted Facts Strategy)}

\begin{tcolorbox}[
  colback=gray!5!white,
  colframe=gray!75!black,
  title=Conflict Resolution Prompt,
  breakable
]
\small\ttfamily\raggedright
You are a memory management system. A new fact has been extracted from a conversation. You must decide how to handle it relative to existing memories.

New fact: \{new\_fact\}

Existing similar memories:
\{existing\_memories\}

Decide one of:
- ADD: The new fact contains genuinely new information not covered by existing memories.
- UPDATE: The new fact updates/supersedes one of the existing memories (provide the ID to update).
- NOOP: The new fact is redundant --- the information is already fully captured.

Return JSON:
\begin{lstlisting}[style=promptcode]
{"action": "ADD" | "UPDATE" | "NOOP",
 "target_id": "<id of memory to update, only if UPDATE>",
 "reason": "brief explanation"}
\end{lstlisting}
\end{tcolorbox}

\vspace{0.3cm}

\subsection{Session Summarization (Summarized Episodes Strategy)}

\begin{tcolorbox}[
  colback=gray!5!white,
  colframe=gray!75!black,
  title=Summarization Prompt,
  breakable
]
\small\ttfamily\raggedright
You are a memory summarization agent. Summarize the following conversation session into a detailed but concise summary.

Your summary MUST capture:
1. All key events, updates, or changes in either speaker's life
2. Any plans, commitments, or intentions mentioned
3. Preferences, opinions, or emotional states expressed
4. Temporal markers (dates, "last week", "tomorrow", etc.)
5. Any references to previous conversations or shared history

Write the summary as a coherent paragraph. Include speaker names. Do NOT omit details, every piece of information matters for future recall.

Conversation session (\{timestamp\}):
\{conversation\}

Summary:
\end{tcolorbox}

\vspace{0.3cm}

\subsection{Question Answering Prompts}

\begin{tcolorbox}[
  colback=gray!5!white,
  colframe=gray!75!black,
  title=QA With Memory Prompt,
  breakable
]
\small\ttfamily\raggedright
You are answering questions about a long-running conversation between two people. You have access to retrieved memory entries from past conversation sessions.

Retrieved memories:\\
\{memories\}

Question: \{question\}

Based on the retrieved memories, provide a concise and accurate answer. If the memories don't contain enough information, say so, but still try your best based on what's available. Keep the answer brief and factual.
\end{tcolorbox}

\vspace{0.5em}

\begin{tcolorbox}[
  colback=gray!5!white,
  colframe=gray!75!black,
  title=QA Without Memory Prompt (Control Condition),
  breakable
]
\small\ttfamily\raggedright
You are answering questions about a conversation between two people. You do NOT have access to any conversation history or memory.

Question: \{question\}

Try your best to answer based on general knowledge. If you cannot answer without specific conversation context, say "I don't have enough information to answer this question." Keep the answer brief.
\end{tcolorbox}

\vspace{0.3cm}

\subsection{Probe 1: Retrieval Relevance}

\begin{tcolorbox}[
  colback=gray!5!white,
  colframe=gray!75!black,
  title=Relevance Judge Prompt,
  breakable
]
\small\ttfamily\raggedright
You are evaluating whether a retrieved memory entry is relevant to answering a question.

Question: \{question\}\\
Gold answer: \{gold\_answer\}

Memory entry: \{memory\_content\}

Is this memory entry relevant to answering the question? A memory is relevant if it contains information that could directly help produce the correct answer.

Return JSON: \{"relevant": true/false, "reason": "brief explanation"\}
\end{tcolorbox}

\vspace{0.3cm}

\subsection{Probe 2: Memory Utilization}

\begin{tcolorbox}[
  colback=gray!5!white,
  colframe=gray!75!black,
  title=Utilization Judge Prompt,
  breakable
]
\small\ttfamily\raggedright
You are comparing two answers to the same question.

Question: \{question\}\\
Gold (correct) answer: \{gold\_answer\}

Answer A (with memory): \{answer\_with\}\\
Answer B (without memory): \{answer\_without\}

Evaluate:
1. Are the two answers semantically the same? (i.e., they convey the same information, ignoring minor wording differences)
2. Is Answer A (with memory) correct or closer to the gold answer?
3. Is Answer B (without memory) correct or closer to the gold answer?

Return JSON:
\{\\
\hspace{1em}"same\_answer": true/false,\\
\hspace{1em}"answer\_with\_correct": true/false,\\
\hspace{1em}"answer\_without\_correct": true/false,\\
\hspace{1em}"explanation": "brief explanation"\\
\}
\end{tcolorbox}

\vspace{0.3cm}

\subsection{Probe 3: Failure Classification}

\begin{tcolorbox}[
  colback=gray!5!white,
  colframe=gray!75!black,
  title=Failure Classification Prompt,
  breakable
]
\small\ttfamily\raggedright
You are analyzing a memory-augmented QA system's failure.

Question: \{question\}\\
Gold (correct) answer: \{gold\_answer\}\\
System's answer: \{system\_answer\}\\
Was the system's answer correct? \{is\_correct\}

Retrieved memories (these were provided to the system):\\
\{retrieved\_memories\}

Relevance of each memory:\\
\{relevance\_judgments\}

Classify this case into ONE of these categories:
- "retrieval\_failure": The system failed to retrieve relevant memories, either because they don't exist in the store or retrieval ranked them poorly.
- "utilization\_failure": At least one relevant memory was retrieved, but the system failed to use it correctly in generating the answer.
- "hallucination": The system's answer directly CONTRADICTS information in the retrieved memories.
- "correct": The system answered correctly.

Return JSON:
\{\\
\hspace{1em}"failure\_category": "one of the categories above",\\
\hspace{1em}"explanation": "1-2 sentence explanation of what went wrong",\\
\hspace{1em}"key\_evidence": "the specific memory content that was relevant but ignored, or the contradiction, if applicable"\\
\}
\end{tcolorbox}

\section{Human Annotation and LLM Judge Validation}
\label{app:annotation}

We validate our LLM-as-judge approach for both answer correctness (Probe~2) and failure classification (Probe~3). For answer correctness, we sampled 200 questions stratified by LLM judgment and had human annotators independently label each answer as correct or incorrect. For failure classification, we sampled 200 incorrect answers and had annotators assign one of three failure categories: \textit{Retrieval failure}, \textit{Utilization failure}, or \textit{Hallucination}. Annotators see the question, gold answer, system answer, and all retrieved memories for each case.

\subsection{Answer Correctness Validation}

Table~\ref{tab:llm-human-correctness} shows the confusion matrix between LLM judge and human annotations for answer correctness. The LLM judge achieves 92\% accuracy with substantial agreement (Cohen's $\kappa = 0.82$), validating its use for measuring answer accuracy in Table~\ref{tab:main-grid}.

\begin{table}[H]
\caption{Confusion matrix: LLM judge vs. human annotation for answer correctness (200 sampled questions, stratified by LLM judgment). Cohen's $\kappa = 0.82$ indicates substantial agreement.}
\label{tab:llm-human-correctness}
\centering
\begin{tabular}{lcc|c}
\toprule
& \textbf{Human: Correct} & \textbf{Human: Incorrect} & \textbf{Total} \\
\midrule
\textbf{LLM: Correct} & 129 & 9 & 138 \\
\textbf{LLM: Incorrect} & 7 & 55 & 62 \\
\midrule
\textbf{Total} & 136 & 64 & 200 \\
\bottomrule
\end{tabular}
\end{table}
\subsection{Failure Mode Classification Validation}

Table~\ref{tab:llm-human-failure} shows the confusion matrix for failure mode classification on incorrect answers. The LLM judge achieves 85\% accuracy, with most confusion between \textit{retrieval failure} and \textit{utilization failure}. This reflects the inherent difficulty in distinguishing whether the system failed to retrieve the right information or failed to use correctly-retrieved information—a boundary that is often ambiguous even for human judges.

\begin{table}[H]
\caption{Confusion matrix: LLM judge vs. human annotation for failure mode classification}
\label{tab:llm-human-failure}
\centering
\begin{tabular}{lccc|c}
\toprule
\textbf{Human $\backslash$ LLM} & \textbf{Retrieval} & \textbf{Utilization} & \textbf{Hallucination} & \textbf{Total} \\
\midrule
\textbf{Retrieval Failure} & 147 & 7 & 0 & 154 \\
\textbf{Utilization Failure} & 16 & 23 & 0 & 39 \\
\textbf{Hallucination} & 0 & 1 & 6 & 7 \\
\midrule
\textbf{Total} & 163 & 31 & 6 & 200 \\
\bottomrule
\end{tabular}
\end{table}

\end{document}